\documentclass[aoas]{imsart}

\RequirePackage{amsthm,amsmath,amsfonts,amssymb}
\RequirePackage{natbib}
\RequirePackage[colorlinks,citecolor=blue,urlcolor=blue]{hyperref}
\RequirePackage{graphicx}
	\usepackage{amsmath}
	\usepackage{graphicx}
	\usepackage{enumerate}
	\usepackage{natbib} 
	\usepackage{url} 
	\usepackage{bm}
\usepackage{subcaption}
\usepackage{graphicx}
\usepackage{multirow}
\usepackage{algorithm}
\usepackage{algorithmicx}
\usepackage{algpseudocode}
\usepackage{amsmath}
\usepackage{caption}
\usepackage{subcaption}
\usepackage{amsmath, amsthm, amssymb}
\usepackage{amsmath}

\newtheorem{prop}{Proposition}
\usepackage{hyperref}

\usepackage{booktabs}

\usepackage{tikz-cd}

\usepackage{amsmath,amssymb,amsthm,mathrsfs,amsfonts,dsfont}

\theoremstyle{theorem} 
\theoremstyle{definition} \newtheorem{definition}{Definition}
\theoremstyle{definition} 
\theoremstyle{remark} 
\theoremstyle{remark} 
\theoremstyle{theorem} 

\startlocaldefs

\endlocaldefs

\begin{document}

\begin{frontmatter}
\title{Renewing Iterative Self-labeling Domain Adaptation with Application to the Spine Motion Prediction}
\runtitle{Renewing Iterative Self-labeling Domain Adaptation}

\begin{aug}
\author{\fnms{Gecheng} \snm{Chen}\ead[label=e1]{gechengchen@tamu.edu}},
\author{\fnms{Yu} \snm{Zhou}\ead[label=e2]{yuznick96@tamu.edu}},
\author{\fnms{Xudong} \snm{Zhang}\ead[label=e3]{xudongzhang@tamu.edu}}
\and
\author{\fnms{Rui} \snm{Tuo}\ead[label=e4]{ruituo@tamu.edu}}

\end{aug}

\begin{abstract}
The area of transfer learning comprises supervised machine learning methods that cope with the issue when the training and testing data have different input feature spaces or distributions. In this work, we propose a novel transfer learning algorithm called Renewing Iterative Self-labeling Domain Adaptation (Re-ISDA). This work is motivated by a cervical spine motion prediction problem, with the goal of predicting the cervical spine motion of different subjects using the measurements of ``exterior features''. The joint distribution of the exterior features for each subject may vary with one’s distinct BMI, sex and other characteristics; the sample size of this problem is limited due to the high experimental cost. The proposed method is well suited for transfer learning problems with limited training samples. The learning problem is formulated as a dynamic programming model and the latter is then solved by an efficient greedy algorithm. Numerical studies show that the proposed method outperforms prevailing transfer learning methods. The proposed method also achieves high prediction accuracy for the cervical spine motion problem.
\end{abstract}

\begin{keyword}
\kwd{Transfer learning}
\kwd{iterative learning}
\kwd{dynamic programming}
\kwd{cervical spine}
\end{keyword}

\end{frontmatter}

\section{Introduction}
A primary task of supervised learning is to learn the input-output relationship within a training set. This learned relationship is described by a model, which can then be used to make predictions for another set of interest, e.g., the testing set. Traditional supervised machine learning algorithms learn models based only on the training data. In practice, the marginal distribution of the training inputs often differs from that of the testing set. See \cite{pan2009survey} for an example in web-document classification. When this difference is large, traditional supervised machine learning algorithms may not give accurate predictions on the testing set, even if their models fitting on the training set are good. In such a circumstance, one needs to consider a transfer learning method, which builds machine learning models based on not only the training data but also the inputs of the testing set.

This work is motivated by an application in studies of human motion and biomechanics. An important topic in this area is to use human body exterior features (e.g., contours, surface markers) acquired by computer vision or other measurement means to determine the underlying physics or mechanics. Studies often focus on one body region of interest in a particular type of physical act (e.g., static vs dynamic, high-speed vs low-speed, sports vs vocational).  The current study is on the cervical spine motion, a critical piece of information for understanding as well as prevention of neck pain and injury. There are two fundamentally different approaches to the measurement of cervical spine motion: 1) surface-based approach, which is most widely used and more practical but is subject to significant error sources such as soft tissue motion artifact in estimating the underlying skeletal (i.e., rigid-body) movement; 2) dynamic radiography, which provides the “gold standard” measurement of skeletal movement but is costly, not well accessible and predisposes human subjects to radiation exposure. In view of the practical challenges in acquiring the dynamic radiography data, we hope to build a subject-specific machine learning model that can predict subjects’ skeletal motion (output) based only on their surface-based measurement (input). In this supervised learning problem, we have the labeled data of only six subjects due to time and budget constraints. We choose five of the subjects as the training set to build the model and the remaining as the testing set. The joint distribution of the exterior features for each subject may vary with one’s BMI, sex and other characteristics. Therefore, we must take the difference into account when constructing surrogate models. In other words, we need to address two major challenges:
\begin{enumerate}
    \item The constructed surrogate models should perform well in spite of the difference between joint distributions of inputs of the training and testing sets;
    \item The size of the labeled samples is limited.
\end{enumerate}

A supervised learning approach accounting for and coping with the discrepancy between the distributions of the training and the testing inputs, e.g., the distinct distribution of each subject’s exterior features, is usually called a domain adaptation (transductive transfer learning) method \citep{pan2009survey, redko2019advances}. 
In domain adaptation, we call the training set the \textit{source domain} and the testing set the \textit{target domain}. Although several methods are available in the domain adaptation literature, they cannot fully address the stated challenges. These existing methods mainly fall into two general categories. The first category, called the instance-based transfer learning \citep{pan2009survey}, proceeds by re-sampling the training data according to suitably chosen weights to approximate the distribution of testing samples. These methods are not suitable for the problems with limited labeled samples, such as the spine motion prediction problem, because the information loss due to the undersampling is more severe in small-data problems. The second category, called the feature-based transfer learning \citep{pan2009survey}, proceeds by finding transformations between the inputs of the training and testing sets so that their distributions become closer. However, how to decide a suitable transformation can be difficult and domain-dependent \citep{pan2009survey}. In addition, we find that the typical methods of this category do not perform well in the spine motion prediction problem; see Section \ref{spine}.

In this article, we adopt the general idea of the iterative self-labeling domain adaptation (ISDA) method \citep{habrard2013iterative} and propose a novel method, called the renewing self-labeling domain adaptation (Re-ISDA), to address the two aforementioned challenges. The ISDA method is a domain adaptation method for classification problems. In this work, we propose a new framework, under which a version of the ISDA method is applicable for regression problems, by formulating the regression problem as a dynamic programming model with uniformly stable algorithms \citep{bousquet2002stability}. We also provide a greedy algorithm to solve this dynamic programming model efficiently, which proceeds by renewing all labels of testing samples labeled in former steps. This renewing step helps avoid a potential issue of the ISDA that the possible mis-labeled samples by a weak classifier in the initial stage of the iterative learning can cause serious harm to the subsequent learning process, which is also called mis-labeling issue \citep{wang2020unsupervised,zhang2017joint}. 

In our synthetic numerical examples, the proposed method outperforms three prevailing domain adaptation approaches. In the spine motion prediction problem, we find that the performance of Re-ISDA is superior to the three prevailing domain adaptation approaches. These numerical results suggest that the proposed Re-ISDA method is highly promising for similar human skeletal motion prediction problems.

The remainder of this paper is organized as follows. We first review the main existing methods of domain adaptation in Section \ref{background}. Then we establish a new theoretical foundation for ISDA and propose the Re-ISDA in Section \ref{Re-ISDA}. A numerical example is investigated in Section \ref{simulation}. In Section \ref{spine} we study the cervical spine motion prediction problem. Concluding remarks and discussion are given in Section \ref{conclusion}.

\section{Background}\label{background}

In this section we will introduce the problem formulation and a brief review for domain adaptation. Also we will explain the relation between domain adaptation and our motivating study. 

\subsection{Problem formulation}\label{problem formulation}

We start with a general formulation of domain adaptation, as discussed in \cite{pan2009survey}. A domain is defined as $D=\{X,P(X)\}$, where $X$ stands for a feature space and $P(X)$ denotes the marginal distribution of these features. Besides, a task $T$ based on a specific domain $D$ is defined as $T = \{Y,P(Y|X)\}$ where $Y$ is the label space and $P(Y|X)$ denotes the conditional distribution. Then a domain adaptation task can be described as \citep{redko2019advances, pan2009survey}: Given a source domain (training set) $D_s$ and learning task $T_s$ based on $D_s$, domain adaptation aims to help improve the learning of the target predictive function on target domain (testing set) $D_t$ where $X_s = X_t$, $P_s(X) \not= P_t(X)$ and $T_s = T_t$.

In this work, we focus on regression problems in domain adaptation. Assume that the source domain $D_s=\{(x^1_s,y^1_s),(x^2_s,y^2_s),...,(x^q_s,y^q_s)\}$ and the target domain as $D_t=\{x^0_t,x^1_t,...,x^{p-1}_t\}$. Besides, choose a labeled sample as a calibration point $(x_c,y_c)$. In Section \ref{implementation detail} we will discuss how to choose this calibration point in detail. Let $(X_s,Y_s)$ and $(X_t,Y_t)$ be the joint distributions of inputs and outputs on the source and the target domains. We suppose that $P(X_s)\neq P(X_t)$, i.e., the two data sets have different input distributions. We also assume that the relationship between the input and output, characterized by the conditional distributions, are the same, i.e., $P(Y_s|X_s)=P(Y_t|X_t)$. Let $(x_s,y_s)$ and $(x_t,y_t)$ be data from the source and the target distributions, respectively. Suppose we are given a loss function $L(h;x,y)$, such as the quadratic loss $L(h;x,y)=(y-h(x))^2$. The observed data are independent copies of $(x_s,y_s)$ and independent copies of $x_t$. The problem of interest is to estimate 
$$h_T:=\operatorname{argmin}_{h}\mathbb{E}L(h;x_t,y_t).$$

\subsection{Existing domain adaptation methods}
We review three kinds of domain adaptation methods in this section: instance based transfer learning, feature based transfer learning and iterative self-labeling domain adaptation.
\subsubsection{Instanced based methods}
Instanced based methods, such as \textit{kernel mean matching} KMM \citep{huang2007correcting}, aim to reduce the distance between the marginal distributions of source and target domain through re-weighting the source samples based on the following formula
\begin{eqnarray}
    R_t^l(f)=E_{(x,y) \in D_s}\frac{P(X_s,Y_s)}{P(X_t,Y_t)}l(f(x),y),
\end{eqnarray}
where $R_t^l(f)$ is the empirical loss of predictive function $f(\cdot)$ on $D_t$, $l(\cdot,\cdot)$ is a loss function, $P(X_s,Y_s)$ and $P(X_t,Y_t)$ are the joint distributions of the source and target domains, respectively. In other words, if we can estimate the weight $\frac{P(X_s,Y_s)}{P(X_t,Y_t)}$ and re-weight the source samples, we can obtain the empirical loss of $f(\cdot)$ on target domain. However, these methods are not suitable for the problems with limited labeled samples, such as the spine motion prediction problem, as this re-weighting does not fully utilize the information of the source samples. we refer to \cite{dai2007boosting}, \cite{chattopadhyay2012multisource}, \cite{daume2009frustratingly}, \cite{duan2012domain} for related discussions.

\subsubsection{Feature based methods}
Feature based methods proceed by finding the optimal transformation between the inputs of the training and testing set to reduce the distance between the marginal distributions of the inputs of training and testing set. An prevailing method, \textit{transfer component analysis} (TCA) \citep{pan2010domain}, chooses a kernel function $\phi(\cdot)$ to map the input of the source and target domains into the reproducing kernel Hilbert space generated by $\phi(\cdot)$, denoted by $\mathcal{H}$, and then minimize the distance between these two marginal distributions, i.e.,
\begin{eqnarray}
    \min_{\phi}\left\|\frac{1}{n_s}\sum_{i=1}^{n_s}\phi(x^i_s)-\frac{1}{n_t}\sum_{j=1}^{n_t}\phi(x^j_t)\right\|_{\mathcal{H}},
\end{eqnarray}
where $x^i_s \in D_s$, $x^j_t \in D_t$, $n_s$ and $n_t$ are the size of source and target domain respectively. In our numerical studies, we find that TCA does not perform well in the spine motion prediction problem; see Sections \ref{simulation}-\ref{spine}. 

\subsubsection{Iterative self-labeling domain adaptation}

\begin{figure}[ht]
    \centering
    \includegraphics[width=\textwidth]{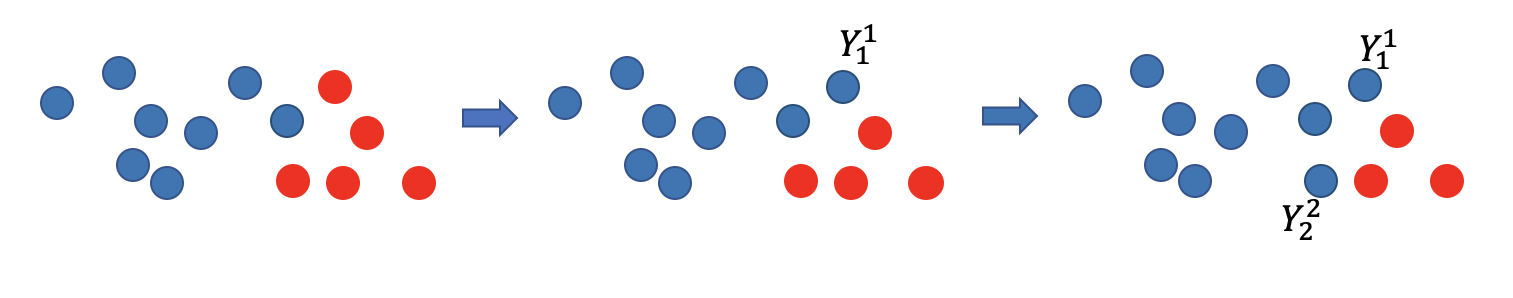}
    \caption{Iteration process of ISDA.}
    \label{fig:ISDA}
\end{figure}

\cite{habrard2013iterative} propose the ISDA based on the domain adaptation SVM (DASVM) \citep{bruzzone2009domain} and gives a theoretical foundation for this domain adaptation classification algorithm. The procedure of ISDA is illustrated in Figure \ref{fig:ISDA}: assume we have some blue points as the training set (source domain) and our mission is to predict the labels of red points (target domain) as shown in the first plot of Figure \ref{fig:ISDA}. In order to predict the labels for target samples more precisely, ISDA mainly repeats these three steps in its iteration process:
\begin{itemize}
    \item First, learn a model $f_n$ from the current labeled and pseudo labeled sample set $S_n$. As shown in the first plot of Figure \ref{fig:ISDA}, it learns a classifier through the labeled samples (blue points);
    \item Second, pseudo-label some target samples based on $f_n$. If we assume it pseudo-labels one target sample each step, one of the target samples will be labeled as $Y_1^1$ as shown in the second plot of Figure \ref{fig:ISDA};
    \item Then incorporate the newly labeled target sample(s) into the source samples to build up $S_{n+1}$ to progressively modify the current classifier. As shown in the third plot of Figure \ref{fig:ISDA}, the ISDA will train a new classifier based on the source samples and the newly labeled target sample, and then label a new target sample using the new classifier as $Y_2^2$.
\end{itemize}
Note that in original ISDA, $k$ source samples will be eliminated from the source domain at each step based on their distance from the classification boundary. In this paper we assume $k=0$.

Habrard proves that the final classifier will work better for domain adaptation problem if all the classifiers during the iteration work better than random guessing and at least one of these classifier is good enough. However, the possible mis-labeled samples by a weak classifier in the initial stage of the iterative learning can cause serious harm to the subsequent learning process, which is also called mis-labeling issue \citep{wang2020unsupervised,zhang2017joint}.

\subsection{Domain adaptation in the spine motion prediction problem}\label{Domain adaptation in our study}

In the spine motion prediction problem, the joint distribution of the surface-based measurement (input features) for each subject may vary with one's distinct BMI, sex and other characteristics (i.e., $P_s(X) \not= P_t(X)$). At the same time, we assume the relationship between the surface-based measurement and the skeletal movement for all subjects, characterized by the conditional distributions, are the same (i.e., $P(Y_s|X_s)=P(Y_t|X_t)$). Based on the above two assumptions, we can formulate the spine motion prediction problem as a domain adaptation problem.


We try the TCA, KMM and ISDA for our spine motion prediction problem and find that although ISDA gets lower prediction error than TCA and KMM, the precision of ISDA remains unsatisfactory, see Section \ref{spine}. In view of this, we propose a new algorithm to enhance the prediction performance of ISDA.

\section{Renewing Iterative Self-labeling Domain adaptation}\label{Re-ISDA}

The proposed Renewing Iterative Self-labeling Domain Adaptation (Re-ISDA) method also follows the general steps of ISDA but it renews all the pseudo labels of target samples at each step to modify their possible error. The proposed Re-ISDA proceeds by iteratively repeating the following three steps:

\begin{enumerate}
    \item Learn a model $f_n$ from the current labeled and pseudo labeled sample set $S_n$;
    \item Pseudo-label some new target samples and update all the pseudo labels given in previous steps based on $f_n$;
    \item Combine the newly labeled target sample(s) and the target samples with updated labels with the source samples to build up $S_{n+1}$. 
\end{enumerate}

\begin{figure}[ht]
    \centering
    \includegraphics[width=\textwidth]{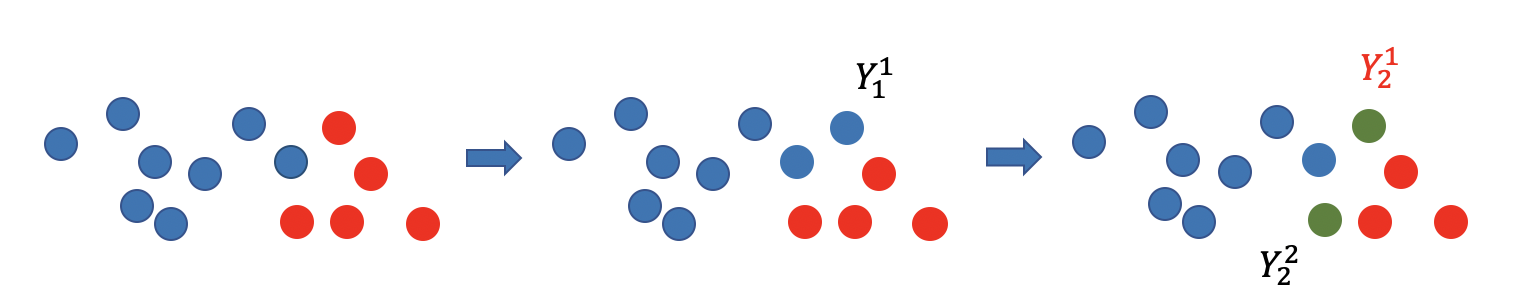}
    \caption{Iteration process of Re-ISDA.}
    \label{fig:reISDA}
\end{figure}

The main steps of Re-ISDA are illustrated in Figure \ref{fig:reISDA}. First, we learn a predictor through the source samples (blue points). In Figure \ref{fig:reISDA}, we pseudo-label only one target sample per step. Then the Re-ISDA labels one of the target samples as $Y_1^1$ which is shown in the second plot of Figure \ref{fig:reISDA}. Next, as shown in the third plot of Figure \ref{fig:reISDA}, the ISDA will train a new predictor using the source samples and the newly labeled target sample with its label $Y_1^1$. Based on this new predictor, it will label a new target sample using the new predictor as $Y_2^2$ and update the pseudo-label $Y_1^1$ by $Y_2^1$ which means the label for the first target sample in the second iteration. By adding this updating strategy, the Re-ISDA has the ability to modify the possible mis-labels in the previous steps and prevent the mis-labeling issue. It is worth noting that in practice, we may pseudo-label multiple target samples in one step; see Section \ref{implementation detail} for more discussions.

In Section \ref{methodology}, we introduce a theoretical foundation of the Re-ISDA by constructing a dynamic programming framework for the domain adaptation regression problems. Some implementation guidelines for the proposed method are provided in Section \ref{implementation detail}. 
\subsection{The proposed methodology}\label{methodology}

The theoretical foundation of ISDA given by \cite{habrard2013iterative} is not applicable in the current context, because it focuses on classification problems but we are interested in regression problems. Before introducing our dynamic programming framework, let us introduce some basic concepts. \cite{bousquet2002stability} define the \textit{uniform stable algorithms} as follows:
\begin{definition}
\textit{An algorithm A has a uniform stability $\beta$ with respect to the loss function $l$ if} $\forall S \in \{X\times Y\}^m, \forall i \in {1,2,...,m}$
\begin{eqnarray}
    \sup_{(x,y) \in S} |l(h_S(x),y)-l(h_{S^{\backslash i}}(x),y)|<\beta,
\end{eqnarray}
\textit{where hypothesis $h_S$ is learned from $S$ and $S^{\backslash i}$ is obtained from $S$ with deleting the $i$-th sample.} 
\end{definition}
It is also shown that $\beta$ decreases at the order of magnitude $O(1/m)$ if the algorithm is stable, where $m$ is the size of $S$. So we can denote it as $\beta_S$ \citep{bousquet2002stability}. According to this definition, we can get an equivalent proposition:

\begin{prop}\label{prop2}

Assume that there exists a function $f$ with uniform stability $\beta_S$ which can fit a big enough labeled set $S=\{(x_1,y_1),(x_2,y_2),...,(x_q,y_q)\}$, then for each point $(x,y)\in S$,
\begin{eqnarray}\label{point_lose}
     \sum_{f\in\mathcal{F}} |f(x)-y|<n\beta_0,
\end{eqnarray}
where $\mathcal{F}= \{f=T(s)| s \subseteq S\backslash (x,y) \}$, $f=T(s)$ means $f$ is trained based on the set $s$, the size of every $s$ is big enough, $n$ is the number of such $f$, $\beta_0=max\{\beta_s|s\subseteq S\backslash (x,y)\}$. 
\end{prop}

We only consider the big enough subsets of $S$ in Proposition \ref{prop2} because when the size of $s$ is too small, the corresponding $\beta$ could be arbitrarily large, which makes the left hand side of (\ref{point_lose}) intractable. If we define the left hand of (\ref{point_lose}) as the \textit{point loss} of a fixed point in $S$, Proposition \ref{prop2} can be interpreted as: if a big enough labeled set can be fitted by a uniformly stable function, then the point loss is small for each fixed point in this set.

In view of Proposition \ref{prop2}, we have an equivalent statement of the original domain adaptation problem. Let $D_{tl}$ be the labeled target domain. The equivalent problem is to find labels $Y_t$ for $D_t$ such that $D_{tl}\bigcup D_s$ can be fitted by a uniformly stable function, i.e., find labels $Y_t$ to minimize the point loss of a calibration point $(x_c,y_c)$, that is, 
\begin{eqnarray}\label{th:original_objective_1}
    \min _{Y_t} \sum_{f\in \mathcal{F}} |f(x_c)-y_c|,
\end{eqnarray}
where $\mathcal{F}=  \{f=T(s)| s \subseteq D_s\bigcup D_{tl}\}$, $D_{tl}$ is the labeled target domain by $Y_t$, the size of $s$ is big enough. But the size of $\mathcal{F}$ in (\ref{th:original_objective}) may be too large, which makes this problem intractable. To address this challenge, we propose an approximation scheme. We define a proper subset of $\mathcal{F}$, denoted as $F$. Then instead of solving (\ref{th:original_objective_1}), we consider the approximated problem
\begin{eqnarray}\label{th:original_objective}
    \min _{Y_t} \sum_{f\in F} |f(x_c)-y_c|.
\end{eqnarray}

\begin{figure}[ht]
    \centering
    \includegraphics[width=\textwidth]{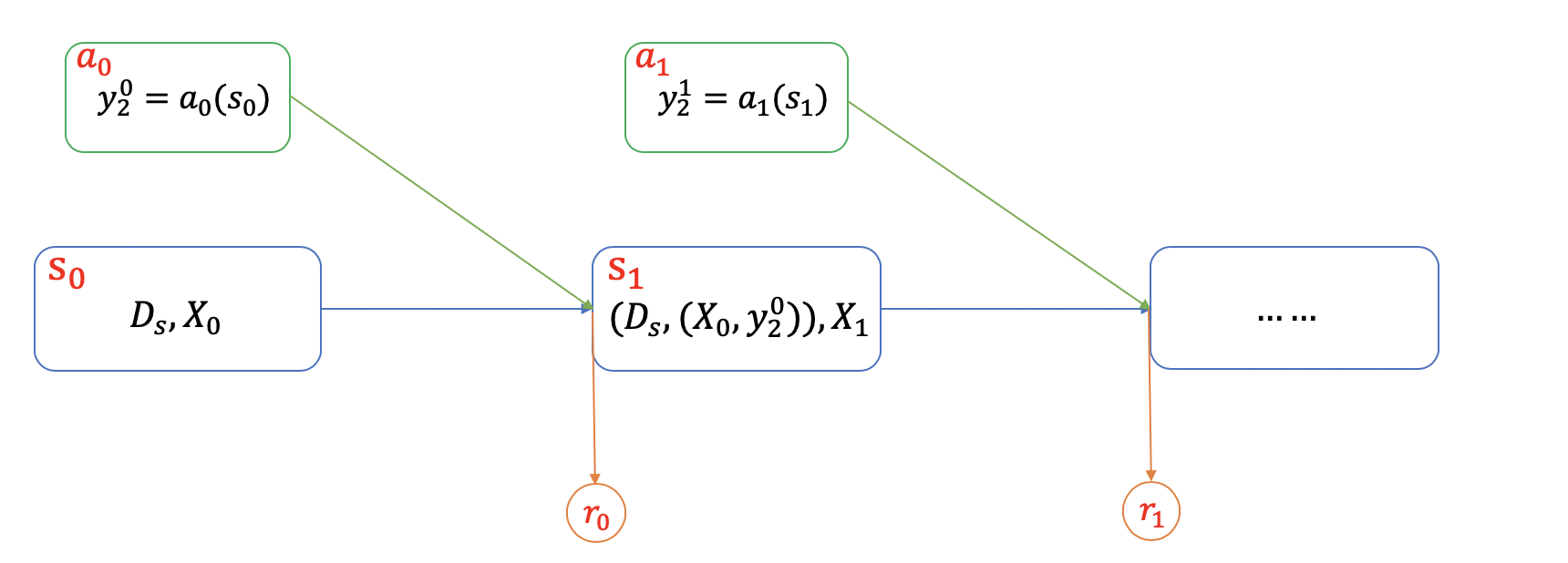}
    \caption{First two states of the system.}
    \label{fig:markov chain}
\end{figure}

Next, we fix the order of the target samples and model the problem (\ref{th:original_objective}) as a dynamic programming problem. First, define state $s_n=\{u_1^n,u_2^n\}$, where $u_1$ is the set of labeled or pseudo-labeled samples, and $u_2$ is the set of $\eta$ newly unlabeled target samples in each state. Also we define action function as $a_n$, $y_2^n$ determined by $a_n$ is the set of pseudo labels for $u_2^n$. The loss at each state caused by the action $a_n$ is denoted as
\begin{eqnarray}\label{lose}
    r(s_n,a_n)=|f_n(x_c)-y_c|,
\end{eqnarray}
where $f_n = T(u_1^n \bigcup (u_2^n,y_2^n))$. Figure \ref{fig:markov chain} shows the first two states of the whole system, where $X_n=\{x_{n\eta},x_{n\eta +1},...,x_{(n+1)\eta-1}\},n=0,1,...,p/\eta-1$. For example, at state $S_0$, $u_1^0$ represents $D_s$, $u_2^0$ stands for $X_0$, pseudo labels $y_2^0$ for $X_0$ is determined by action $a_0$ and the loss $r(s_0,a_0)=|f_0(x_c)-y_c|$, where $f_0 = T(D_s \bigcup (X_0,y_2^0))$.

Our goal is to find the optimal policy $\{a_n\},n=0,1,...,p/\eta-1,$ that minimizes the sum of loss, i.e.,
\begin{eqnarray}\label{th:new_obj1}
    \min _{\{a_n\}} \sum_{i=0}^n r(s_n,a_n).
\end{eqnarray}

Now we consider problem (\ref{th:original_objective}) on the set $F := \{f_0,f_1,...,f_n\}$, $n=0,1,\ldots,p/\eta-1 $. Then the problem is equivalent to the dynamic programming problem (\ref{th:new_obj1}).

\begin{figure}[ht]
    \centering
    \includegraphics[width=\textwidth]{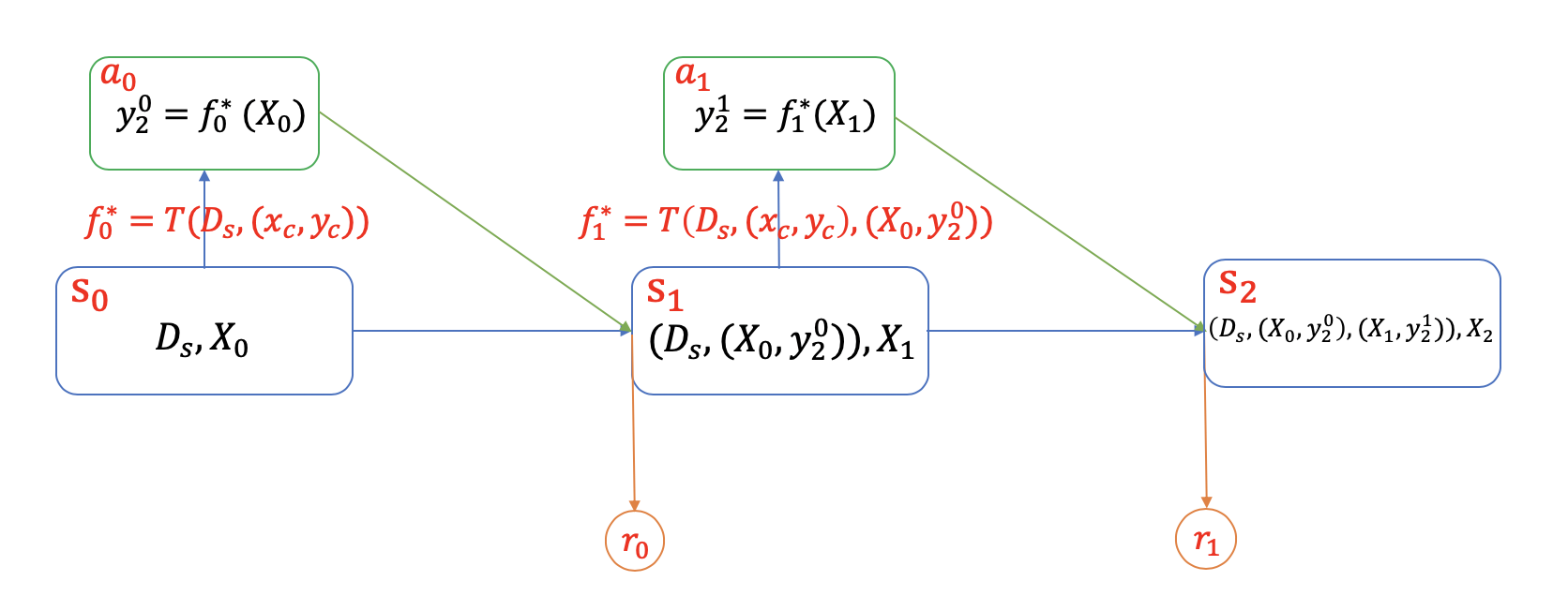}
    \caption{Iteration process for action function (\ref{action1}).}
    \label{fig:function1}
\end{figure}

Solving (\ref{th:new_obj1}) exactly is usually still computationally intractable. Thus it is reasonable to consider approximate algorithms to tackle (\ref{th:new_obj1}).
The ISDA can be regarded an simple greedy algorithm to cope with (\ref{th:new_obj1}), that is, to solve
\begin{eqnarray}\label{th:new_obj}
    \min_{a_n} r(s_n,a_n),
\end{eqnarray}
step by step, for $n=0,1,\ldots,p/\eta-1$ if we assume the action function as follows
\begin{eqnarray}\label{action1}
     y_2^n=f_n^*(u_2^n),
\end{eqnarray}
where $f_n^* = T(u_1^n \bigcup (x_c,y_c))$ and $T$ is a stable basic learner. This function means if we need to find suitable labels for $u_2^n$ to minimize the loss in (\ref{lose}) we can just use the predictor trained by $\{u_1^n \bigcup (x_c,y_c)\}$ to predict the labels. And the solution to the previous step is assumed as known and fixed in the subsequent steps. The detailed process can be seen as Figure \ref{fig:function1}. At the sate $S_0$, the pseudo label for $X_0$ is given by the action $a_0$ where the predictor $f_0^*$ is trained from $D_s \bigcup (x_c,y_c)$, i.e., $f_n^* = T(u_1^n \bigcup (x_c,y_c))$. At the same time, the action $a_0$ will result in a loss $r_0$. Then the iteration continues.

As a simple greedy algorithm, the ISDA has the following deficiency: once a pseudo labeling error occurs, it can never be corrected. If the errors at early iterations are large, the entire downstream process can be highly disturbed, which is also called mis-labeling issue \citep{wang2020unsupervised,zhang2017joint}. This phenomenon will be shown in our experiments.

We propose a modification of the ISDA action function, which enables a self-correcting capability of the method and stabilizes the iteration process.
Let $u_1^n = \{D_s,d^n\}$, where $d^n=(d_t^{n-1},y_{pl}^{n-1})$, $d_t^{n-1}$ is the set of target samples added in former steps, and $y_{pl}^{n-1}$ is the set of their pseudo labels given by former steps. The modified action function consists of two components:
\begin{eqnarray}\label{action2}
     y_2^n=f_n^*(u_2^n),
\end{eqnarray}
and
\begin{eqnarray}\label{action3}
     y_{pl}^n = f_n^*(d_t^{n-1}),
\end{eqnarray}
where $f_n^* = T(u_1^n \bigcup H^t)$ and $T$ is a stable basic learner. In other words, in each epoch, we first pseudo label the newly added target samples, and then renew the pseudo labels of target samples added before. The first step (\ref{action2}), following the ISDA algorithm, explores the labels for the newly added points. The second step (\ref{action3}) updates the former pseudo labels in order to correct any possible errors. Note that (\ref{action3}) will not dramatically change the pseudo labels of former points because the difference between $y_{pl}^n$ and $y_{pl}^{n-1}$ should be small, which promises a stable iteration process. We call this new method \textit{Renewing Iterative Self-labeling Domain Adaptation} (Re-ISDA).

\begin{figure}[ht]
    \centering
    \includegraphics[width=\textwidth]{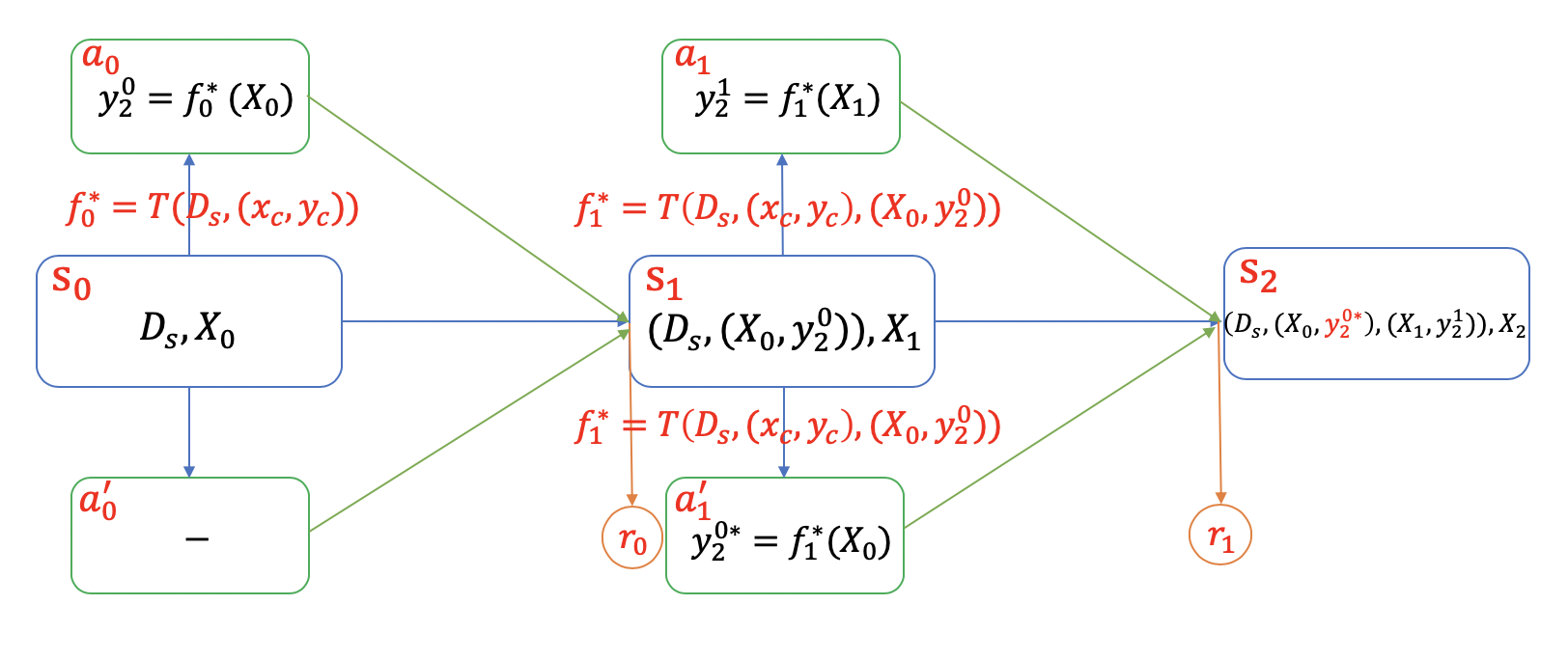}
    \caption{Iteration process for action functions (\ref{action2}) and (\ref{action3}).}
    \label{fig:function2}
\end{figure}

The iteration process of Re-ISDA is illustrated in Figure \ref{fig:function2}. For example, at the state $s_1$ there will be two actions: $a_1$ and $a'_1$: $a_1$ gives the pseudo label of $x_1$ based on the predictor $f_1^*$ trained on $\{D_s \bigcup H^* \bigcup {(X_0,y_2^0)}\}$; $a'_1$ renews the label $y_2^0$ ($y_{pl}^0$) by $y_2^{0*}$ ($y_{pl}^1$) based on $f_1^*$. 

The detailed algorithm of Re-ISDA is shown in Algorithm \ref{alg}. We need to choose a basic learner $Q$, the calibration point $(x_c,y_c)$, the order of adding target samples and the size of added samples $\eta$ at each step when implementing the Re-ISDA. Some suggestions for implementation details of the proposed algorithm will be provided in Section \ref{implementation detail}.
 \begin{algorithm}[htb]
        \caption{Training steps for Re-ISDA}\label{alg}
        \begin{algorithmic}[1]
            \Require
                ordered training set $S=\{(x_{s_1},y_{s_1}), (x_{s_2},y_{s_2}),...,(x_{s_n},y_{s_n})\}$, a labeled calibration point $(x_c,y_c)$, testing samples $X_t=(x_{t_1},x_{t_2},...,x_{t_m})$, the number of newly added testing samples $\eta$, number of epochs $P$, initial iterative set $S_0=S \bigcup (x_c,y_c)$, a basic learner $Q$.
            \Ensure
                Final predictor $\hat{f}_P(\cdot)$
            \For{$p$ in $1:P$}
                \State Train a predictor $\hat{f}_p(\cdot)$ based on $S_{p-1}$ and $Q$
                \State Pseudo-predict set $G_p={\{(x_{t_1},\hat{f}_p(x_{t_1})), (x_{t_2},\hat{f}_p(x_{t_2})),...,(x_{t_{rp}},\hat{f}_p(x_{t_{rp}}))\}}$
                \State Renew iterative set $S_p=S_0 \cup G_p$
            \EndFor
        \end{algorithmic}
    \end{algorithm}
    
\subsection{Implementation details for Re-ISDA}\label{implementation detail}

In this section, we offer some practical guidelines for implementing the proposed Re-ISDA method.

\subsubsection{Basic learner $Q$} 

We recommend using feed-forward neural networks (FNNs) as a basic learner. According to \cite{xu2012robustness}, FNNs are robust learners. Besides, FNNs can handle huge data sets and do not require complex transformation or representation learning for the raw data before training \citep{lecun2015deep}. In order to get a well-performing FNN, one needs to choose suitable hyper-parameters such as the number of layers and nodes, learning rate, the number of training epochs, and so on. Detailed tuning guidelines for FNNs can be found in \cite{schmidhuber2015deep,goodfellow2016deep,deng2014deep}. 

\subsubsection{Calibration points $(x_c,y_c)$}
In domain adaptation problems we may face two different situations: 1) no labeled target samples can be used; 2) a few labeled target samples are available. Here we give some guidelines for choosing the calibration points under different occasions:
\begin{itemize}
    \item If we only have the labels for the source domain, we recommend choosing the source sample as the calibration point which has the least average distance between the target samples;
    \item If we only have one labeled target sample, we can choose this one as the calibration point;
    \item If we have several labeled target samples, we can choose one point randomly from them as the calibration point or implement Re-ISDA for each labeled target sample and ensemble their results to give the final prediction results.  
\end{itemize}

\subsubsection{Order of added target samples}
We make the suggestions below for the order of added target samples under two types of data:
\begin{itemize}
    \item If the data is in a time-series format, we should maintain the original order to keep the dynamic properties.
    \item If the data does not have a temporal structure, we recommend sorting the target samples from small to large according to their distance to the calibration point.
\end{itemize}

\subsubsection{Size of added samples in each step $\eta$}

The numerical experiments in Section \ref{spine} show that the Re-ISDA with smaller $\eta$ will lead to more precise prediction results and a more stable iteration process. 
On the other hand, a smaller $\eta$ will require more iteration steps and thus increase the computational cost. Therefore, we recommend setting $\eta$ as small as possible provided that the computational cost is affordable.

\section{Simulation studies}\label{simulation}

In this experiment we assess the performance of proposed Re-ISDA method and compare the results with TCA, KMM, ISDA and the situation without domain adaptation. We use the Friedman Function \citep{friedman1981projection,friedman1991multivariate} as the test function, which is defined as:
\begin{eqnarray}\label{friedman}
    f(\mathbf{z}) = 10\sin(\pi z_1z_2)+20(z_3-0.5)^2  +10z_4 +5z_5.
\end{eqnarray}
Suppose the source domain is $[0.2,1.2]^5$.
A Halton sequence \citep{halton1964algorithm} with 80 samples in $[0.2,1.2]^5$ and the corresponding labels are used as the training set, denoted as $X_s=\{(x_1,y_1),(x_2,y_2),...,(x_{80},y_{80})\}$. From the target domain we choose 41 samples given by adding $0.2$ to each dimension of the inputs of the first 41 samples in $X_s$, i.e., $X_t=\{x_1+\mathbf{0.2},x_2+\mathbf{0.2},...,x_{41}+\mathbf{0.2}\}$, where $\mathbf{0.2}=(0.2,0.2,0.2,0.2,0.2)^T$. Additionally, the calibration point is $(x_c,y_c)=(x_{1}+\mathbf{0.2}, f(x_{1}+\mathbf{0.2}))$, where $f$ is the function in (\ref{friedman}). Further, we reorder the testing samples from the smallest to the largest by their distance from $(x_c,y_c)$. The performance of each predictor is evaluated by the root mean squared error (RMSE) defined as:
\begin{eqnarray}
    RMSE =\sqrt{\frac{1}{n}\sum_{n}^{i=1}(\hat{y}_i-y_i)^2},
\end{eqnarray}
where $n$ is the size of testing set, $\hat{y}_i$ is the predicted value of $i$-th testing sample and $y_i$ is its true value. The implementation details are given below:
\begin{itemize}
    \item The NN without domain adaptation uses a neural network with structure $c(5,10,5,1)$, i.e., a fully connected neural network with four layers, in which the first layer has five dimensions; the second layer has ten dimensions; and so on. Its learning rate is 0.1, and the number of training epochs is 300. This network is also used as the base learner in the following algorithms;
    \item The KMM method uses the Gaussian kernel function with parameter $0.5$ to re-weight the source samples, and then the NN used above is implemented to learn from the re-weighting set;
    \item The TCA method transforms the source input and target input to a new source input set and a new target input set, respectively, to minimize the distance between their distributions0 in a reproducing kernel Hilbert space. The new source and target input set have five dimensions. After the transformation, the new source set is used to train the base learner mentioned before and then to predict the values of the new target set;
    \item The ISDA pseudo labels two samples in each step according to the new order of target domain, and the predictor learner is the neural network with the same structure as mentioned before;
    \item The Re-ISDA is implemented similarly to ISDA except that it renews labels of all added target samples at the current state.
\end{itemize}

\begin{figure}[ht]
    \centering
    \includegraphics[width=0.7\textwidth]{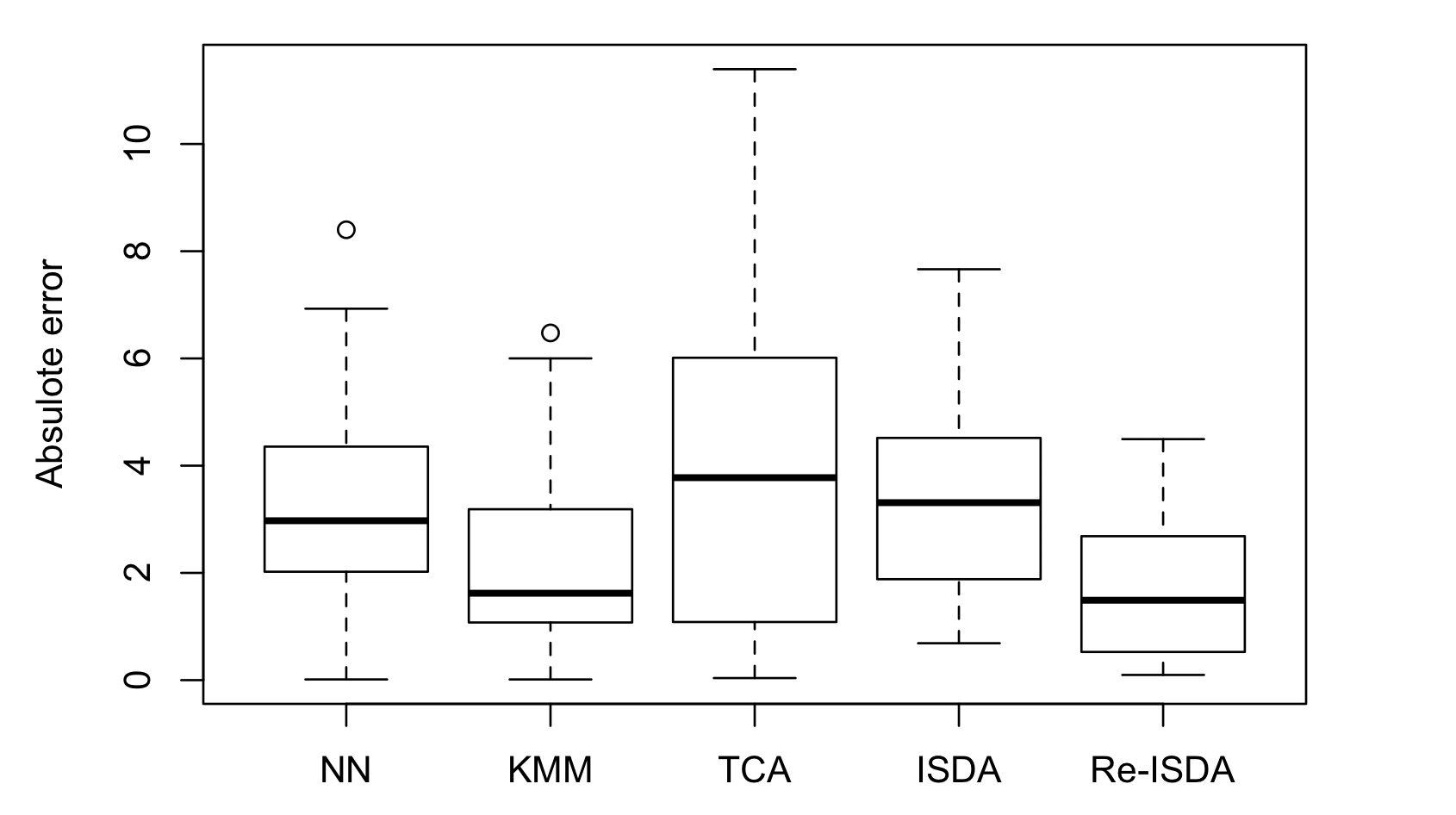}
    \caption{Absolute error of the five methods for Friedman Function.}
    \label{fig:absolute error for friedman}
\end{figure}

The RMSEs of the four methods are given in Table \ref{tab:table1}. Figure \ref{fig:absolute error for friedman} gives the box plot of the absolute error of five methods. Table \ref{tab:table1} and Figure \ref{fig:absolute error for friedman} show that the proposed Re-ISDA has the lowest overall prediction error.

    \begin{table}[ht]
        \centering
        \caption{RMSEs for the five methods on Friedman Function.}
        \begin{tabular}{cccccc}
        \toprule
             & NN without DA & TCA & ISDA & Re-ISDA & KMM \\
        \midrule
             RMSE & 3.623 & 8.901 & 4.070 & 2.033 & 2.70\\
        \bottomrule
        \end{tabular}
        \label{tab:table1}
    \end{table}
    
\begin{figure}[ht]
    \centering
    \includegraphics[width=\textwidth]{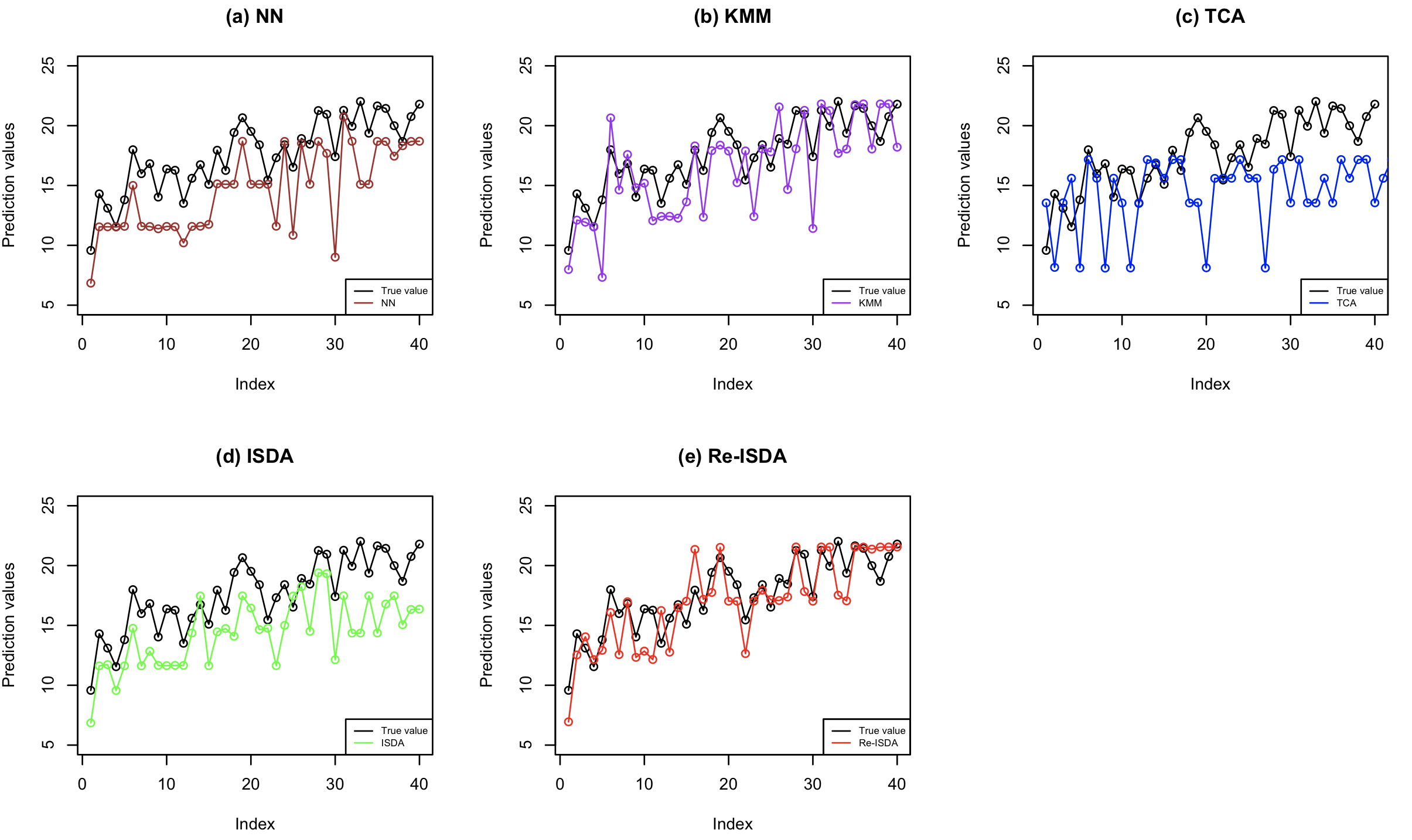}
    \caption{Prediction values of the five methods for Friedman Function.}
    \label{fig:predction for friedman}
\end{figure}

The pointwise predictive value of five methods for every point is shown in Figure \ref{fig:predction for friedman} where the black points represent the true values. We can make the following observations from the Figure \ref{fig:predction for friedman}:
\begin{itemize}
    \item The brown line (NN) is almost all below the black one (true value) and can not follow the changing trend of the true value;
    \item The difference between the purple line (KMM) and the black line is significant, especially for the former half. The possible reason might be the loss of information caused by re-weighting;
    \item The blue line (TCA) vibrates heavily and has large errors for some points. As a result, the performance of TCA is not desirable;
    \item The performance of green line (ISDA) is much better than NN and TCA. However, its performance for the latter half points is much worse than the former half ones. The reason might be the mis-labeling issue \citep{wang2020unsupervised,zhang2017joint};
    \item The red line (Re-ISDA) is distributed much evenly on both sides of the black line compared with the above methods. Additionally, it follows the changing trend of true value much better and has better performance for the latter half points than ISDA. As a result, the Re-ISDA gains the lowest RMSE among the four methods.

\end{itemize}

\section{Cervical spine motion prediction}\label{spine}
In this section, we apply the proposed Re-ISDA method to the cervical spine motion prediction problem. The numerical results show that the Re-ISDA is more suitable for the cervical spine motion prediction problem compared with some prevailing transfer learning methods. 
\subsection{Background}\label{spine_background}

Cervical spine, a highly complex multi-joint structure, supporting the head weight and providing the mobility and flexibility of the head \citep{bogduk2000biomechanics}, is susceptible to injuries that occur acutely and chronically. To design effective injury prevention and control, an accurate characterization
of the cervical spine motion is needed for a better understanding of the neck biomechanics as well as the pathomechanics of neck injury and neck pain \citep{bogduk2000biomechanics,huelke1986cervical,swartz2005cervical}.

Conventional approaches to measuring \textit{in vivo} three-dimensional (3D) skeletal motion and position often involve using optical markers systems and inertial sensors placed on the body surface. Such surface-based measurements are subject to soft tissue motion artifacts and therefore do not reflect the underlying skeletal motion accurately \citep{cappozzo1996position,li2012inaccuracy,tsai2011effects,benoit2006effect}. In addition, marker-based or sensor-based approaches are more suited for measuring the kinematics of long bones (i.e., extremities). For structures such as the cervical spine where relatively small individual bones are inter-connected by complex joints, surface-based measurements are simply too error-prone. Dynamic stereo-radiography (DSX) has proven to be an accurate, effective, and non-invasive method of quantifying
3D \textit{in vivo} cervical spine motion and is regarded as the gold standard 4. However, the technology is not widely accessible and incurs tremendous cost both in development and use.

Previous studies have focused on using surface markers to estimate 3D bone motion or accessing the correlations between measurements from surface markers and from rigid markers (i.e., the markers are mounted on the bones) but are all limited to the lower extremity \citep{cappozzo1997surface,cereatti2006reconstruction,fukaya2012two}. In a recent study, Nichoson et. al developed machine learning algorithms to predict scapular kinematics using optical motion measurement data \citep{nicholson2019machine}, which further inspires
us to use statistical or machine learning approaches to estimate and predict 3D cervical spine motion using optical motion measurements. This and our recent success in measuring cervical spine motion using DSX and optical motion capture inspire us to pursue new machine learning approaches to estimate cervical vertebrae motion.

\subsection{Experimental data acquisition}
\begin{figure}[ht]
    \centering
    \includegraphics[width=\textwidth]{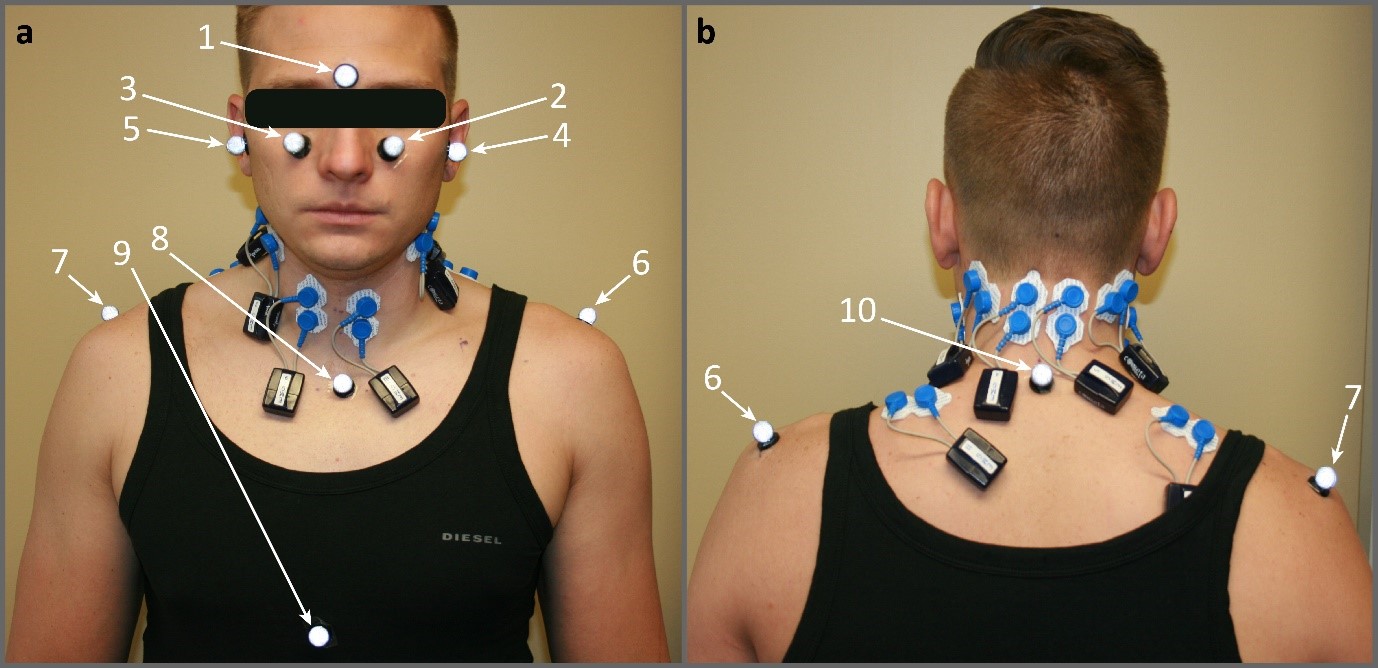}
    \caption{Retro-reflective spherical surface marker placements: a the front view; b the back view.  1 glabella (marker Forehead), 2\&3 inferior border of each orbit on both sides (marker LORBIT and marker RORBIT), 4\&5 the tragion notches of both sides (marker LMP and marker RMP), 6\&7 acromion processes (marker LSHO and marker RSHO), 8 suprasternal notch (marker CLAV), 9 sternum (marker STRN), and 10 C7 spinous process (marker C7).}
    \label{fig:datafigure1}
\end{figure}

In this experiment, participants performed self-paced head-neck flexion-extension tasks (i.e., dynamic free range-of-motion tasks) with the head-neck moving primarily in the sagittal plane while their cervical spine region was imaged continuously using a DSX system \citep{zhou2020state}. A 12-camera Vicon motion capture system (Vantage-Series, Vicon Motion Labs, Oxford, UK) was used to monitor and record the head-neck motions from retro-reflective spherical surface markers placed on ten anatomical landmarks: glabellae, inferior border of each orbit (left and right), the tragion notches (left and right), acromion processes (left and right), suprasternal notch, sternum, and C7 spinous process as shown in Figure. \ref{fig:datafigure1}.

The recorded trajectories of the surface markers were tracked in a global laboratory coordinate system, resulting in time-histories of three coordinates (x, y, and z) in each marker.  

Three-dimensional (3D) orientation and position of each vertebra were determined via a previously validated volumetric model-based tracking algorithm \citep{anderst2011validation,bey2006validation,anderst2009validation,anderst2013six}.  Cervical sagittal spinal curvature was measured at each time frame using the Cobb method \citep{cobb1948outline,harrison2001reliability,vrtovec2009review} where the Cobb angle was defined by the angle between the two inferior endplates of the vertebrae.  However, due to the lack of the endplate in C1 (the topmost vertebra), the cervical sagittal spinal curvature in this study was calculated by the Cobb angle between C2 and C7 inferior endplates as shown in Figure \ref{fig:datafigure2}. The raw data consists of the following parts:
\begin{figure}[ht]
    \centering
    \includegraphics[width=0.6\textwidth]{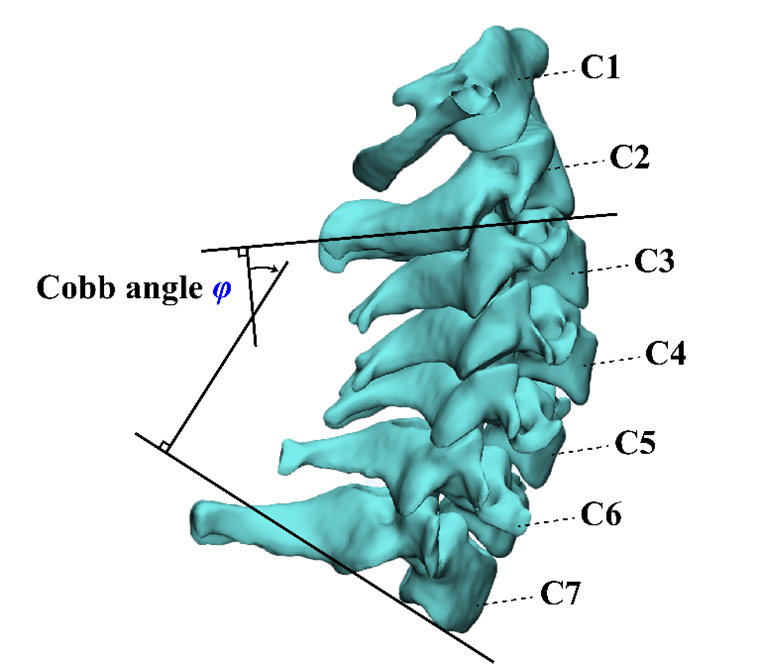}
    \caption{Cervical spinal curvature determined by Cobb angle.}
    \label{fig:datafigure2}
\end{figure}

\begin{itemize}
    
    \item The coordinates information given by 9 markers which named as C7, CLAV, LMP, LORBIT, LSHO, RMP, RORBIT, RSHO, STRN ($9 \times 3$ in total);
    \item BMI information and C7-MMP statistic which is the distance between marker C7 and the midpoint of marker RMP and marker LMP calculated at the neutral static posture;
    \item Neck angle, measured as the angle between the horizontal plane and the plane defined by the marker C7, LMP, and RMP; 
    \item Head angle, measured as the angle between Franfort plane and the horizontal plane, where the Frankfort was defined by the surface markers LMP, RMP, LORBIT, and RORBIT;
    \item C7-MMP, the distance between the marker C7 and the midpoint of marker RMP and marker LMP at each frame during the dynamic head-neck flexion-extention motion trial;
    \item Cobb angle as measured by dynamic biplane radiography.
\end{itemize}

\subsection{Pre-processing}

We shall use the marker coordinates information, BMI information, C7-MMP statistic, neck angle, head angle, and C7-MMP as the input data to predict the corresponding Cobb angle. In contrast to the BMI information and C7-MMP, which do not evolve in time, the rest of the input data are time series. They capture the position of the corresponding markers and are collected sequentially with fixed time intervals. 

As a pre-processing step, we take the difference of every two consecutive frames of the time-series data. We believe that this difference can be used to enhance the prediction performance because it reflects the velocity of each marker. We include this computed ``velocity information'' of the time-series input variables as part of the input data for each candidate learning method. In summary, the input dimension is 62.

\subsection{Objective and problem formulation}

We have labeled information (both marker information and its corresponding Cobb angle information) of six people: subject \#3, 4, 17, 31, 34, 38. The goal is to identify the suitable method to predict the Cobb angle information based on the corresponding marker information, which is required and necessary for a better understanding of the neck biomechanics. To this end, we partition the dataset into two parts: 
\begin{itemize}
    \item Source domain (training set): subject \#3, 4, 31, 34, 38 (called $D_{s1},D_{s2},D_{s3},D_{s4},D_{s5}$);
    \item Target domain (testing set): subject \#17 (called $D_t$).
\end{itemize}
Additionally, we assume that a labeled target sample $\{x^*,y^*\}$ is available (working as the calibration point), which is the first sample of subject \#17 with its true label. This assumption is reasonable in real-world applications because we only need one frame of the biplane radiography data, which can be collected via a conventional static radiography equipment in an inexpensive experiment. 

In this work, we consider four candidates of domain adaption methods: the proposed Re-ISDA and other three prevailing transfer learning methods. The performance of each method is measured by the RMSE under the testing data.

\subsection{Data analysis}
\begin{figure}[ht]
    \centering
    \includegraphics[width=0.5\textwidth]{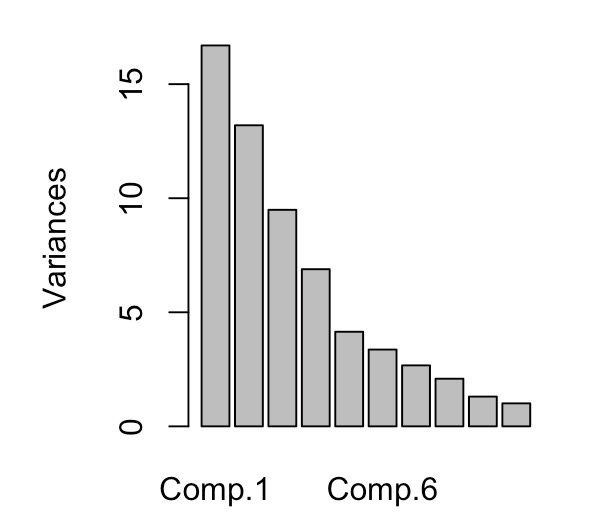}
    \caption{The variance of components in PCA.}
    \label{fig:PCA}
\end{figure}

First, we apply the max-min normalization on both the source and target domain data set to stabilize the forthcoming training process. Next, we employ the principal component analysis (PCA) to reduce the dimensions. Figure \ref {fig:PCA} shows the variances of the first ten components given by PCA. The eleventh principal component accounts for only about 1\% of the total variance, which is almost negligible. In this experiment, we choose the first ten components as the new design.
\begin{figure}[ht]
    \centering
    \includegraphics[width=\textwidth]{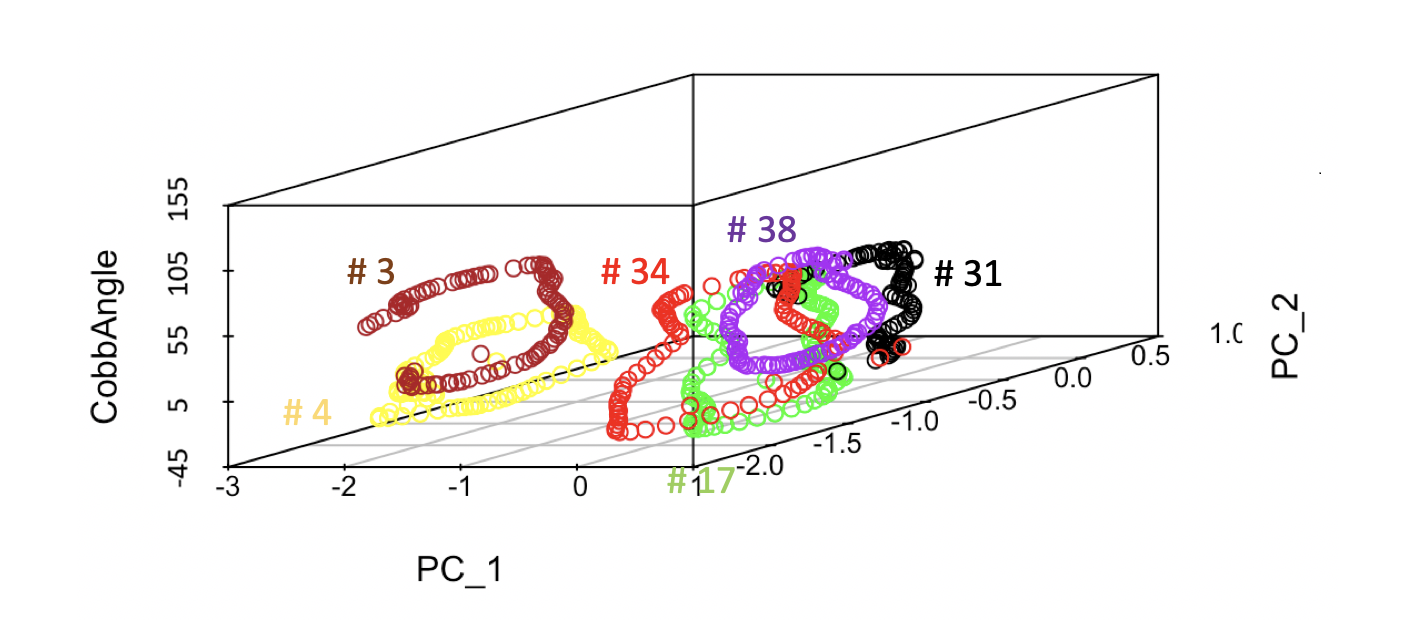}
    \caption{The distributions of first two components of \#3, 4, 31, 34, 38, 17 against their Cobb angle (in degrees).}
    \label{fig:rawDist}
\end{figure}
Figure \ref{fig:rawDist} shows the distribution of the first two components of \#3, 4, 31, 34, 38, 17 against their true Cobb angle information, which can approximately reflect their distribution in the 63 dimensions space. From Figure \ref{fig:rawDist}, the data from each subject can be fitted by a separate curve. Thus, we consider the task of prediction as a multi-source transform learning problem. Denote the function trained by each source as $f_i=T(D_{si}),i=1,2,3,4,5$, and the prediction function for target domain $D_t$ as $f_t$. 
For convenience, we choose $f_t$ according to the following rule:
\begin{eqnarray}
    f_t = \arg \min_{f_{i}}|f_{i}(x^*)-y^*|,
\end{eqnarray}
In this experiment, we choose $f_5$ as $f_t$ and the initial source domain $D_s = D_{s5}$.

Then we implement the four methods as follow:
\begin{itemize}
    \item the NN without domain adaptation uses a neural network with structure $c(10,16,16,8,1)$. Its learning rate is 0.1, and the number of training epochs is 1500. This network is also used as the base learner in the following algorithms;
    \item the KMM method uses Gaussian kernel function with parameter $2$ to re-weight the source samples, and the NN used above is implemented to learn from the re-weighting set;
    \item the TCA method skips the PCA, and is implemented directly on the original input with 63 dimensions. The new source and target input set have ten dimensions. After the transformation, the new source set is used to train the base learner mentioned before and then to predict the values of the new target set;
    \item the ISDA pseudo labels two samples in each step according to the new order of target domain, and the predictor learner is the neural network with the same structure as mentioned before;
    \item the Re-ISDA is implemented similarly to ISDA except that it renews labels of all added target samples at the current state. 
\end{itemize}

\begin{figure}[ht]
    \centering
    \includegraphics[width=0.7\textwidth]{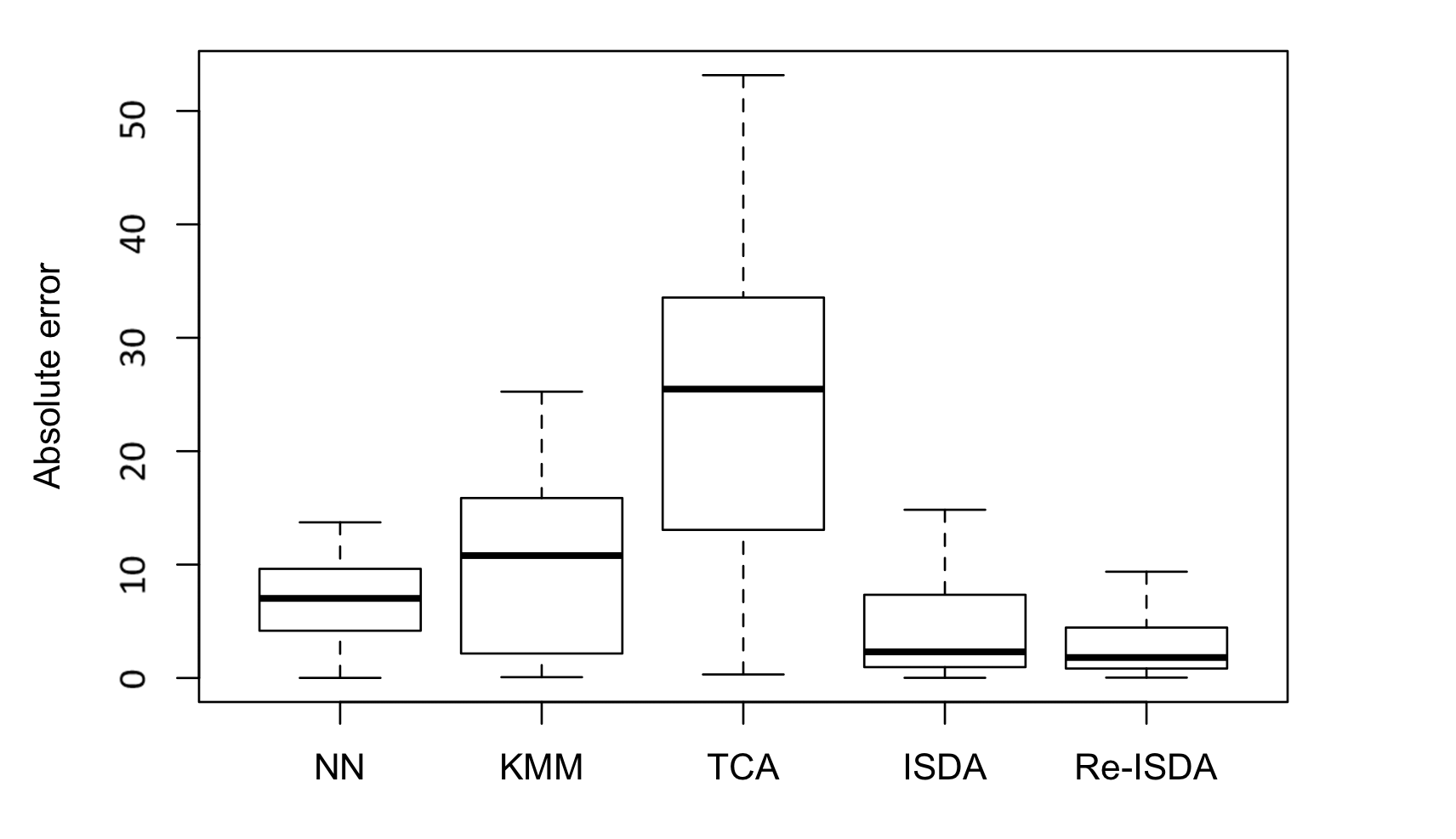}
        \caption{Absolute error of the five methods in the spine motion prediction problem.}
    \label{fig:absolute error for Neck}
\end{figure}

The RMSEs of the four methods are given in Table \ref{tab:table2} and Figure \ref{fig:absolute error for Neck} shows the absolute error of the five methods in the box plot. These two figures prove that the Re-ISDA has the best performance compared with the other four methods. 
    \begin{table}[ht]
        \centering
        \caption{RMSEs for the five methods in the spine motion prediction problem}
        \begin{tabular}{cccccc}
        \toprule
             & NN without DA & TCA & ISDA & Re-ISDA  & KMM \\
        \midrule
             RMSE & 8.19 & 28.88 & 6.21 & 2.68 & 13.2\\
        \bottomrule
        \end{tabular}
        \label{tab:table2}
    \end{table}
\begin{figure}[ht]
    \centering
    \includegraphics[width=\textwidth]{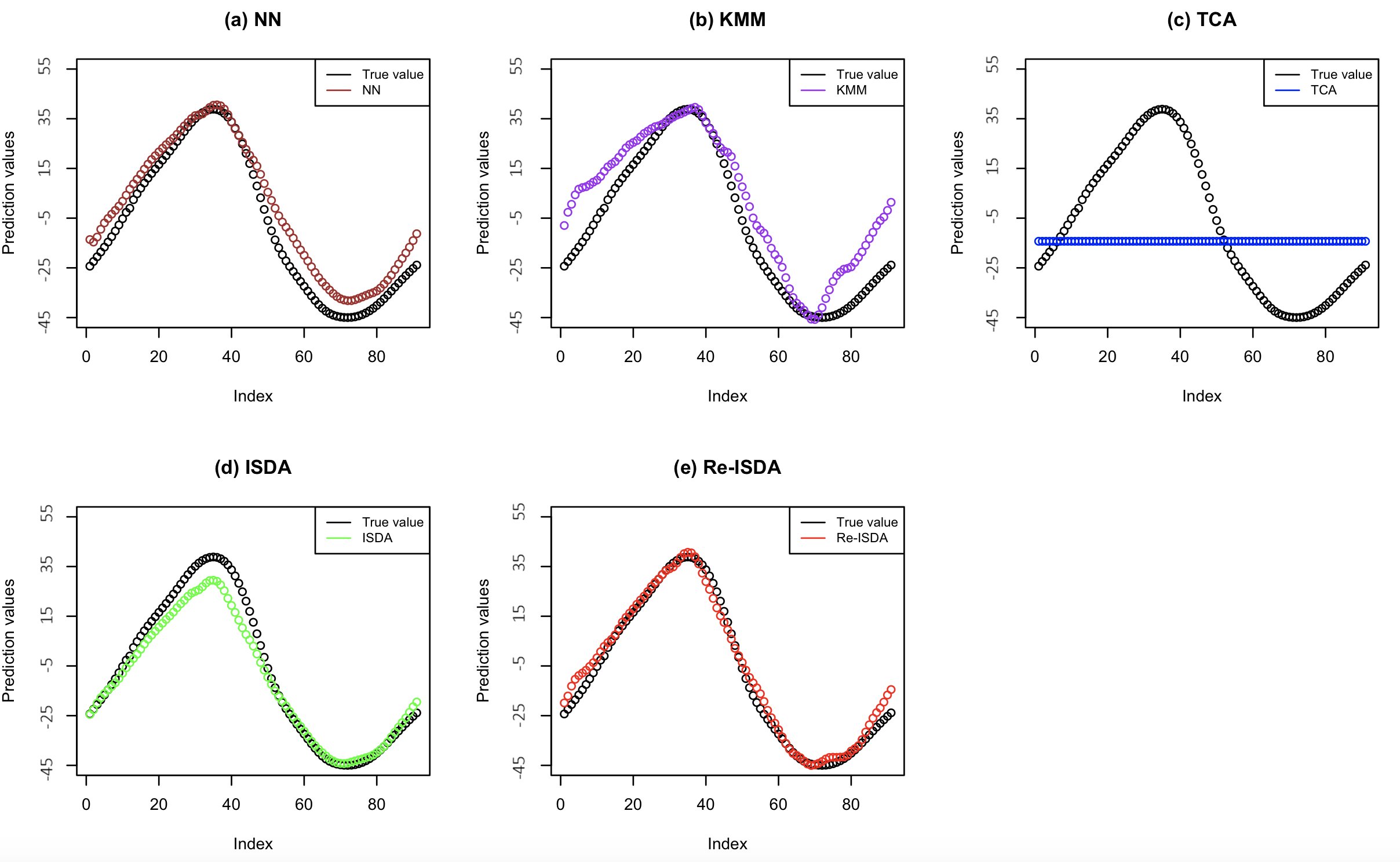}
    \caption{Prediction values of the five methods in the spine motion prediction problem.}
    \label{fig:predction for Neck}
\end{figure}

The pointwise predictive value for every point is shown in Fig.\ref{fig:predction for Neck} where the black points represent the true values. Clearly, we can see from the Fig.\ref{fig:predction for Neck}:
\begin{itemize}
    \item The brown line (NN) deviates far away from the black one (true value), especially for the latter half, representing that the traditional predictor can not work well for this situation;
    \item The performance of the purple line (KMM) is worse than NN. In this experiment, the difference between weights of different source samples is up to $10^{10}$. As a result, many points with tiny weights are actually deleted from the training set, which causes the under-fitting of the model;
    \item The blue line (TCA) is approximately a straight line. We have tried different kernel functions and their parameters for the TCA but get some straight lines just like the line shown here. The possible reason might be that the TCA makes the source domain harder to learn for a regression mission;
    \item The green line (ISDA) works much better than the above three methods and is very close to the black line. However, its performance for the former half part is not desirable. The possible reason for it might be the little flexibility of ISDA so that it can not modify the labels of the former samples based on the points coming afterward;
    \item The red line (Re-ISDA) is approximately the same as the black one, which indicates that our method works highly well. This result shows that the proposed Re-ISDA method can transfer the predictor through different marginal distributions and is flexible enough to modify the pseudo labels in the former steps.
\end{itemize}
    
\begin{figure}[ht]
    \centering
    \includegraphics[width=0.7\textwidth]{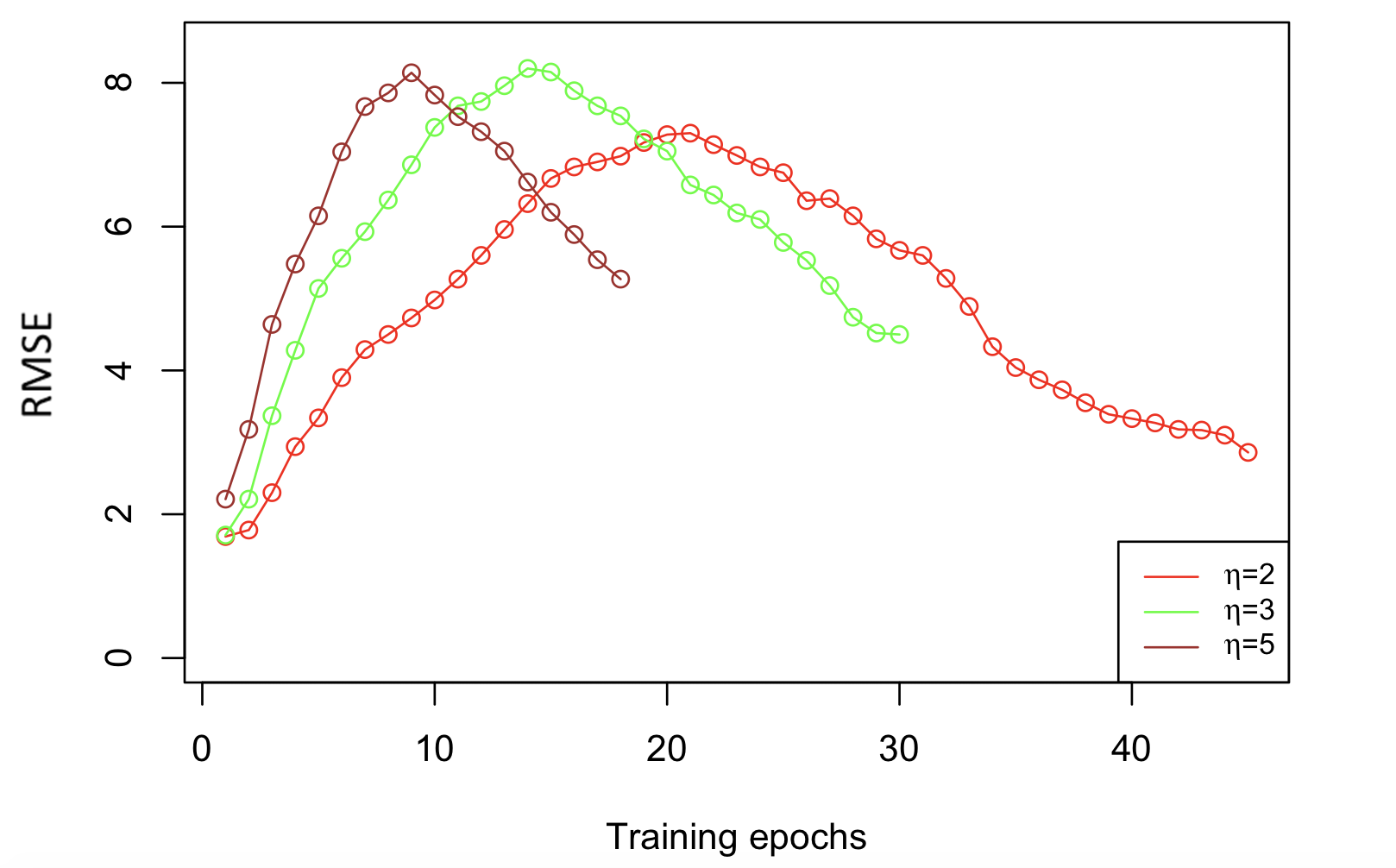}
    \caption{Performance of Re-ISDA for different $\eta$ during their training processes.}
    \label{fig:eta}
\end{figure}

Additionally, we test the performance of Re-ISDA for different $\eta$. Figure \ref{fig:eta} shows the RMSEs for the labeled points (both newly labeled and labeled in previous steps) at each step of Re-ISDA when $\eta=2,3,5$. In the iteration processes, the RMSEs of predictors increase at the beginning and then fall, proving that the Re-ISDA modifies the prediction error of the added points through the new coming samples. Besides, if we choose the smaller $\eta$, the performance of the predictor changes more smoothly and ends at a lower RMSE. However,  a smaller $\eta$ means more training epochs and higher computation cost, so that one needs to make a trade-off between precision and cost.

\section{Conclusions and Discussion}\label{conclusion}

This work proposes a novel method called Re-ISDA for the domain adaptation problem with limited labeled samples based on the ISDA algorithm. We first formulate the learning problem as a dynamic programming model and then propose a greedy algorithm to solve it efficiently. The proposed method updates all labels of the target samples (both newly added and added in former iterations) at each iteration, leading to higher flexibility and fewer mis-labeling issues than ISDA. Our numerical experiments and a cervical spine motion prediction example confirm the prediction performance of the proposed method.

The computational cost of the proposed method is slightly higher than the other domain adaptation methods we have compared with.
Thus the proposed method should be more suitable for moderate sized data. Besides, the Re-ISDA is more suitable for the time-series data because it can incorporate the time-series information.  In this work, we only consider prediction problems, but the proposed method, with proper modifications, should also be applicable to classification problems.


\begin{funding}
Zhang's work was supported by the Centers for Disease Control and Prevention/National Institute for Occupational Safety and Health under a research grant R010H010587. Tuo's work was supported by NSF under grants DMS-1914636 and CCF-1934904.
\end{funding}



\bibliographystyle{imsart-nameyear} 
\bibliography{ref}       

\end{document}